# 三维地理场景下带引线点要素注记优化配置的 Beams 移位方法

魏智威[1]，杨乃[2]，许文嘉[3]，丁愫[4]，李敏敏[1]*，李游[1]，郭仁忠[1,5]

[1] 人工智能与数字经济广东省实验室（深圳）广东 深圳 518132
[2] 中国地质大学（武汉）地理与信息工程学院 湖北 武汉，430074
[3] 北京邮电大学信息与通信工程学院，北京 100876
[4] 浙江农林大学环境与资源学院，浙江 杭州 311300
[5] 深圳大学建筑与城市规划学院智慧城市研究院，广东 深圳 518060
(liminmin@gml.ac.cn)

**摘　要:** 在三维地理场景中，通过为点要素添加带引线的注记可以显著提高目标的可辨识性。引线注记在位置配置上具有较大的自由度，但现有方法多以有限位置候选模型为基础，既没有有效利用图面空间，也难以顾及注记间相对关系。因此，本文面向二维屏幕显示需要，将引线注记的动态配置过程建模为地图目标移位问题，以三角网表达注记间空间关系，基于注记配置约束计算注记受力，应用经典的 Beams 移位方法迭代计算注记新位置。实验结果表明，依据本文方法生成的配置结果能有效避免注记压盖，且邻近注记间平均方向偏差较小；另外，该方法也适用于不同引线类型的注记配置。本文也讨论了提高注记配置效率的分块处理策略和分析了不同邻近图构建策略对实验结果的影响。

**关键词：** 地理场景；三维可视化；注记配置；地图综合；移位算法

## Optimized 3D Point Labeling with Leaders Using the Beams Displacement Method

Wei Zhiwei[1], Yang Nai[2] Xu Wenjia[3], Ding Su[4], Li Minmin[1]*, Li You[1], Guo Renzhong[1, 5]

[1] *Guangdong Laboratory of Artificial Intelligence and Digital Economy, Guangdong Shenzhen* 518132
[2] *School of Geography and Information Engineering, China University of Geosciences (Wuhan), Hubei Wuhan* 430078
[3] *School of Information and Communication Engineering, Beijing University of Posts and Telecommunications, Beijing* 100876
[4] *College of Environmental and Resource Science, Zhejiang Hangzhou* 311300
[5] *Research Institute for Smart City, School of Architecture and Urban Planning, Shenzhen University, Guangdong Shenzhen* 518060

**Abstract:** In three-dimensional geographical scenes, adding labels with leader lines to point features can significantly improve their visibility. Leadered labels have a large degree of freedom in position configuration, but existing methods are mostly based on limited position candidate models, which not only fail to effectively utilize the map space but also make it difficult to consider the relative relationships between labels. Therefore, we conceptualize the dynamic configuration process of computing label positions as akin to solving a map displacement problem. We use a triangulated graph to delineate spatial relationships among labels, and calculate the forces exerted on labels considering the constraints associated with point feature

---

labels. Then we use the Beams Displacement Method to iteratively calculate new positions for the labels. Our experimental outcomes demonstrate that this method effectively mitigates label overlay issues while maintaining minimal average directional deviation between adjacent labels. Furthermore, this method is adaptable to various types of leader line labels. Meanwhile, we also discuss the block processing strategy to improve the efficiency of label configuration and analyze the impact of different proximity graphs.

## 1 引 言

随着信息通信技术，特别是人工智能、虚拟/增强现实等新技术的发展，地图表达形式逐渐从静态信息描述扩展为多维动态空间信息可视化[1]。其中，注记通过文字传递目标语义信息精准直观，是地图语言的重要组成，也是三维可视化表达的核心要素[2]。

依据注记要素数据类型，可划分为点、线、面和体；考虑到实际应用通常是面向屏幕空间进行显示，体在标注时亦可作为特殊的面或点考虑[3]。其中，点要素注记配置因其求解的复杂性，一直是现有研究的重点，相关工作可分为局部优化和全局优化。局部优化方法主要关注单个或局部注记的配置，强调结合注记配置规则量化注记位置得分，并依据得分依次确定注记的最优位置。如文献[4]和[5]综合考虑压盖、位置优先级和位置关联性等结合模拟退火方法确定注记位置；文献[6]综合考虑位置优先级、清晰易读性等结合禁忌搜索算法确定注记位置；文献[7]则基于备选位置与背景要素的压盖程度进行打分确定注记位置，文献[8]和[9]则基于要素压盖结构量化得分确定注记位置。局部优化方法具有配置简单、效率高等优势，但易陷入局部最优，有时无法获取满足要求的配置结果。全局优化方法则是利用遗传算法、蚁群算法等全局优化方法将所有点要素视为整体进行解算，如文献[10]和[11]考虑位置优先级和注记冲突分别结合遗传算法和蚁群算法实现了注记配置。考虑到遗传算法可能存在局部收敛问题，文献[12]将粒子群算法和遗传算法结合实现注记配置，文献[13]则将遗传算法和禁忌搜索算法结合实现注记配置。但是，随着要素数量的增多，全局优化的组合会呈爆炸式增长，导致求解效率低且质量不佳。故后续部分学者考虑结合点要素空间分布特征指导注记配置以提高效率，如采用聚类方法将整体划分为多个子群求解[14,15]，结合注记分布密度规划注记配置次序[16,17]等。

随着三维应用的普及，学者也尝试结合三维可视化特点将上述方法扩展至三维场景，如文献[18]考虑三维场景近大远小等特点将遗传算法用于三维场景中点要素注记的配置；文献[19]和[20]则考虑三维场景语义结构，将启发式方法用于建筑物注记的配置。但是，上述方法多要求注记与对应点要素紧密邻接，在视角频繁变换的三维场景中易产生指代对象不明等问题，故部分学者考虑利用引线连接点要素与注记以提高指代对象的一致性，如文献[21]提出了一种尺度自适应的带引线点要素注记配置方法，有效实现了三维地形场景下的地名标注；文献[22]则结合动态规划策略提出了适宜于建筑物的带引线注记标注方法。另外，文献[23]对比了三维室内场景下带引线注记和纹理贴附式注记的优劣，证明了带引线注记能更明确和完整地表达目标信息。但是，上述带引线的点要素注记配置方法仍多以有限位置候选模型为基础，既没有有效利用图面空间，也难以顾及注记间相对关系。实际上，由于注记与对应点要素以引线连接，注记在位置配置上相比已有方法具有更大的自由度[24]，有必要针对三维地理场景下带引线点要素注记设计合适的配置方法以提高注记配置质量。

由上文分析可知，三维地理场景下带引线点要素注记配置的核心是在考虑三维地理场景特点的同时，通过移动注记获取满足配置约束的结果。移位是地图综合领域的基础操作，已发展有较多成熟的地图要素移位方法，如适宜面要素移位的 Beams 算法，其能在移动面要素的同时保持要素间空间关

系[25, 26]。考虑到三维地理场景目前仍多面向二维屏幕进行显示，若将二维屏幕上的注记以文字占据的绑定矩形代替，就能将注记的配置问题建模为面要素移位问题；同时，若利用 Beams 移位算法移位相比已有方法就能有效保持要素间空间关系。故本研究试图将三维地理场景下带引线点要素注记的配置过程建模为面要素移位问题，设计相应的注记位置优化配置 Beams 方法。

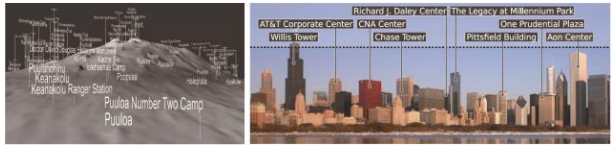

图 1.带引线点要素注记示例。(a)夏威夷岛地形数据标注结果[21]；(b)芝加哥市天际线标注结果[22]。

## 2. 问题定义

### 2.1 问题描述

本文针对三维地理场景中带引线点要素注记研究其位置配置方法。其中，直线是目前普遍采用的引线形式，依据直线引线的方向和引线与注记连接点的性质，可以将其划分为四类：(1)引线方向固定，连接点位置固定，如图 2(a)；(2)引线方向不固定，连接点位置固定，如图 2(b)；(3)引线方向不固定，连接点位置不固定，如图 2(c)；(4) 引线方向固定，连接点位置不固定，如图 2(d)。其中，若引线方向不固定易造成图面混乱；而连接点位置固定则会减小注记位置配置的自由度，引线方向固定连接点位置不固定(类型 4)则是实际应用中较为广泛的类型，如图 1 所示。本文亦针对类型 4 注记研究其配置方法；同时，本文方法亦可扩展至另外三类注记的配置，详见章节 5.2。

本文问题可定义如下：给定二维屏幕上待显示的点要素集合 $P=\{p_1, p_2, ..., p_m\}$，依据引线参数(用户设置的长度、方向和连接点)和要素标注字段可确定注记的初始位置和内容，点要素与对应注记以引线连接；注记可用其绑定矩形表示，记为 $R=\{r_1, r_2, ..., r_m\}$，点要素与其注记的引线连接点集合记为 $P'=\{p'_1, p'_2, ..., p'_m\}$，点 $p_m$ 与注记的引线即为 $p_m p'_m$，移动 $R$ 中的注记并更新 $P'$ 以获取满足约

束条件的配置结果，约束条件定义见章节 2.2。

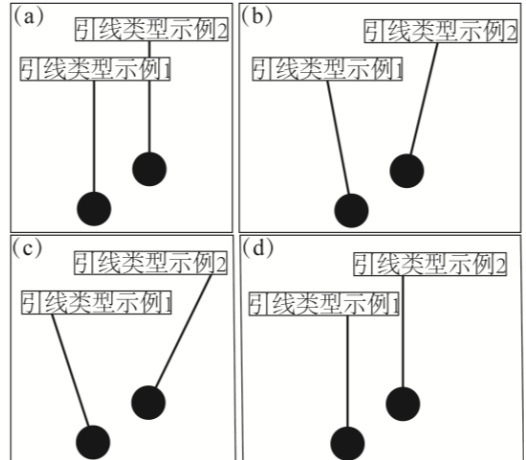

图 2.引线注记分类。(a)引线方向固定，连接点位置固定；(b)引线方向不固定，连接点位置固定；(c)引线方向不固定，连接点位置不固定；(d)引线方向固定，连接点位置不固定。

### 2.2 配置约束定义

三维地理场景中的注记配置在考虑三维场景表达特殊性的同时，也必须遵循相应的可视化原则。本文参考文献[18]，[23]和[25]，总结了三维地理场景中带引线点要素注记配置的表达需求。需要说明的是，本文基于二维屏幕显示需要计算注记配置结果，约束、计算过程和评价指标均为基于屏幕显示需要定义。

**(1)几何约束**

位置：在固定引线方向的前提下，注记的位置不能横向偏离对应点要素距离过大，否则无法保证引线连接点与注记连接；

尺寸：注记尺寸需符合三维场景近大远小的表达要求。

**(2) 关系**

距离关系：为保证三维场景的可读性，注记与注记间或注记与已有其它图面要素(如点要素)间的距离应大于一定阈值 $d_{min}$，例如，在距离目标 30 公分的情况下，人眼能分辨的最小间隔距离为 $0.2mm$[2]。

拓扑关系：在屏幕上显示的注记与注记间或注记与点要素间不能产生压盖；

空间方向关系：注记间的空间方向关系应尽量反映注记对应点要素间的空间方向关系。

**(3) 分布**

注记移动总距离：注记的总移动距离应尽可能小。

需要说明的是三维地理场景复杂多样，实际应用中用户也可依据需要调整或扩展配置约束，如可能会考虑场景的结构特征等进行配置[3,19,24]。

## 3. 方　法

Beams 移位方法多用于面状要素的移位，其原理是将面要素用邻近图关联，并将邻近图类比材料力学中的杆件结构，面要素由于距离过近或压盖而产生冲突需要移动(即对杆件结构施加力)，杆件因力的作用发生形变或位移而产生能量，通过求解能量最小化问题即可计算杆件移位和变形后的最优形状和位置，Beams 移位方法的求解过程详见文献[25]。因此，利用 Beams 移位方法进行注记配置需要：(1)依据注记配置的需要生成初始结果；(2)将注记对应的绑定矩形关联为邻近图；(3)依据章节 2.2 的注记配置约束计算邻近图受力；(4)基于构建的邻近图及其受力，利用 Beams 算法迭代求解获取邻近图中各点的新位置，即优化后的注记位置。

### 3.1　初始注记布局

依据章节 2.1 的问题描述，给定二维屏幕上待显示的点要素集合 $P=\{p_1, p_2,...,p_m\}$，依据用户设置的引线长度、方向、初始引线连接点位置和最大注记字号 $W_{max}$，可生成点要素注记的初始布局，注记对应的绑定矩形记为 $R=\{r_1, r_2,...,r_m\}$，牵引点集合记为 $P'=\{p'_1, p'_2,...,p'_m\}$，如图 3(a)所示。其中，注记的字号由视点到点要素的平面距离决定。假设离视点最近的点为 $p_{nearest}$，其字号即为 $W_{max}$；给定点 $p_m$，视点到 $p_m$ 的平面距离为 $d_m$，则 $p_m$ 注记字号 $W_m$ 计算见公式(1)。

$$W_m = \frac{d_{nearest}}{d_m} \times W_{max} \qquad (1)$$

$d_{nearest}$ 为视点到 $p_{nearest}$ 的平面距离；另外，考虑到人眼对字符的可辨识度，字号需大于一定阈值 $W_{min}$，若 $W_m \leq W_{min}$，则 $W_m = W_{min}$。

### 3.2　邻近图构建

三角网(Delaunay Triangulation，缩写为 DT)以图的形式表达地图目标间邻近关系和空间分布，能有效反映地图目标间相互联系、相互依存和相互影响的能力[27]。因此，本文采用 DT 表达注记间相互关系，如图 3(b)所示。其中，若三角网关联的两个注记距离较远或被其它注记阻隔，注记实际上不邻近，因此，需要删除 DT 中长度大于阈值($T_d$)的边和穿过注记的边。邻近图构建流程如下：

(1) 以注记的绑定矩形中心点集构建 DT，结果见图 3(b)；

(2) 删除 DT 中距离较长的边和穿过注记的边，如图 3(b)中红色边，邻近图生成结果见图 3(c)。

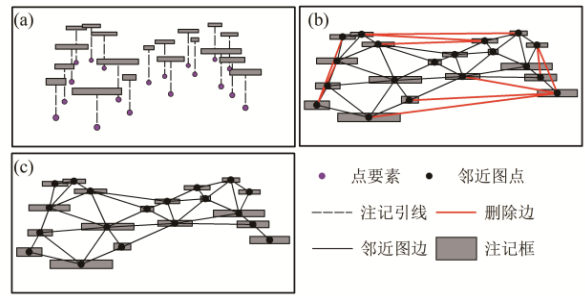

图 3.(a)初始注记配置结果(注记以其绑定矩形替代表示)；(b)DT；(c)最终生成的邻近图。

### 3.3　力的计算

Beams 方法用于面状要素移位时，若面要素之间距离小于阈值或压盖(即发生冲突)会产生力；同理，若注记之间或注记与其它已有图面要素(如点要素或其它已有注记)之间的距离小于阈值或压盖也会发生冲突，即对邻近图产生力。

**(1) 注记之间的斥力计算**

若两个注记相离但距离小于阈值 $d_{min}$，则需要为两个注记施加使它们相离的斥力以保证注记间距离大于 $d_{min}$，计算见公式(2)和(3)。

$$f_l = 0.5 * \frac{\overrightarrow{p_l p_l}}{|\overrightarrow{p_r p_l}|} \times (d_{min} - |\overrightarrow{p_r p_l}|) \qquad (2)$$

$$f_r = 0.5 * \frac{\overrightarrow{p_l p_r}}{|\overrightarrow{p_l p_r}|} \times (d_{min} - |\overrightarrow{p_l p_r}|) \qquad (3)$$

其中，$p_l$ 和 $p_r$ 为左右两注记上相距最近的点，如图 4(a)所示。

若两个注记相交，则需要为两个注记施加使它们相离的斥力以保证注记不相交且注记间距离大于 $d_{min}$。考虑到注记以绑定矩形替代，可以在二维屏幕的 $x-$、$x+$、$y-$ 和 $y+$ 四个方向上分别计算使注记间距离大于 $d_{min}$ 的受力，注记最终受力可依据四个方向上的最小受力计算。假设表示两个注记的绑定

矩形 $r_i$ 和 $r_j$ 分别记为 $[x^i_{min}, y^i_{min}, x^i_{max}, y^i_{max}]$，$[x^j_{min}, y^j_{min}, x^j_{max}, y^j_{max}]$，其中，$x_{min}, y_{min}, x_{max}, y_{max}$ 分别表示绑定矩形在 $x$ 方向和 $y$ 方向上的最小值和最大值，则沿 $x$-、$x$+、$y$-和 $y$+四个方向移动 $r_i$ 使其与 $r_j$ 距离大于 $d_{min}$ 的受力 $f_{x-}$、$f_{x+}$、$f_{y-}$、$f_{y+}$ 计算公式见(4)-(7)，如图 4(b)所示；其中，移动 $r_j$ 使其与 $r_i$ 距离大于 $d_{min}$ 的过程与移动 $r_i$ 使其与 $r_j$ 距离大于 $d_{min}$ 的过程对偶。

$$f_{x-} = x^i_{max} - x^j_{min} + d_{min} \quad (4)$$
$$f_{x+} = x^j_{max} - x^i_{min} + d_{min} \quad (5)$$
$$f_{y-} = y^i_{max} - y^j_{min} + d_{min} \quad (6)$$
$$f_{y+} = y^j_{max} - y^i_{min} + d_{min} \quad (7)$$

基于公式(4)-(7)，两相交注记受力大小分别为 $0.5*\min(|f_{x-}|,|f_{x+}|,|f_{y-}|,|f_{y+}|)$，$r_i$ 的移动方向为力 $\min(|f_{x-}|,|f_{x+}|,|f_{y-}|,|f_{y+}|)$ 方向，$r_j$ 与 $r_i$ 移动方向相反。

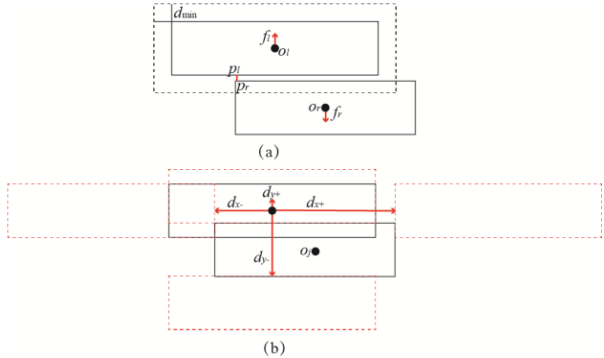

图 4.注记间冲突时力的计算。(a)注记相离但距离小于阈值；(b)注记压盖。

**(2) 注记与已有图面要素间的斥力计算**

若注记与已有图面要素(如点要素或其它已有注记)之间的距离小于阈值或产生压盖，则需要对注记施加远离该要素的斥力。以注记与点要素存在冲突为例，由于点要素无法移动，存在冲突则需要对注记施加远离对应点要素的斥力；但是，若注记受到来自两个相反方向的点要素斥力，则仅靠注记移动无法解决该冲突。因此，为避免上述问题，本文对存在冲突的注记在二维屏幕的 $x$-、$x$+、$y$-和 $y$+四个方向上分别计算可能解决其与对应点要素冲突的斥力，计算见公式(8)-(11)，如图(5)所示。

$$f_{x-} = \frac{\overrightarrow{p_ip_{x-}}}{|\overrightarrow{p_ip_{x-}}|}(|\overrightarrow{p_ip_{x-}}| + d_{min}) \quad (8)$$

$$f_{x+} = \frac{\overrightarrow{p_ip_{x+}}}{|\overrightarrow{p_ip_{x+}}|}(|\overrightarrow{p_ip_{x+}}| + d_{min}) \quad (9)$$

$$f_{y-} = \frac{\overrightarrow{p_ip_{y-}}}{|\overrightarrow{p_ip_{y-}}|}(|\overrightarrow{p_ip_{y-}}| + d_{min}) \quad (10)$$

$$f_{y+} = \frac{\overrightarrow{p_ip_{y+}}}{|\overrightarrow{p_ip_{y+}}|}(|\overrightarrow{p_ip_{y+}}| + d_{min}) \quad (11)$$

其中，$p_i$ 为点要素，$p_{x-}$、$p_{x+}$、$p_{y-}$ 和 $p_{y+}$ 为点 $p_i$ 到注记外包矩形四个方向上的投影点。若注记只与一个点要素冲突，则其受力为 $\min(|f_{x-}|,|f_{x+}|,|f_{y-}|,|f_{y+}|)$；若注记受到来自多个点要素的斥力，则需依据文献[25]中力的合成方法，依次选择来自不同点要素的 1 个同方向(夹角小于等于90°)斥力进行合成，合成后值最小的力即为该注记受力。

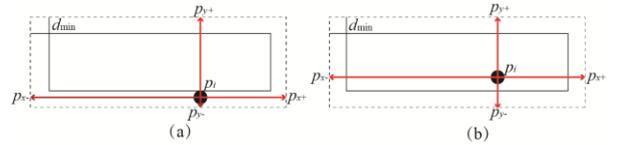

图 5.点要素与注记冲突时力的计算。(a)注记与点要素相离但距离小于阈值；(b)注记与点要素压盖。

**(3) 引线连接点与注记之间的吸引力计算**

由于本文引线方向设置为固定，若注记横向移位距离过大，则无法保证引线与注记连接。因此，若存在注记横向移位距离过大，则需要为其添加一个吸引力 $f_{atr}$ 保证引线与注记相连，计算见公式(12)。

$$f_{atr} = \overrightarrow{p'_ip'_{nearest}} \quad (12)$$

其中，$p'_i$ 为引线连接点，$p'_{nearest}$ 为 $p'_i$ 到对应注记的最近点。

**(4) 屏幕边缘斥力约束**

注记不能超出屏幕边缘，因此，若注记与屏幕边缘间的距离小于距离阈值 $d_{min}$，则屏幕边缘会对注记施加斥力 $f_{screen}$，计算见公式(13)。

$$f_{screen} = \frac{\overrightarrow{p_sp_r}}{|\overrightarrow{p_sp_r}|} \times (d_{min} - |\overrightarrow{p_sp_r}|) \quad (13)$$

其中，$p_s$ 和 $p_l$ 为屏幕与注记上相距最近的两点。

### 3.4 迭代求解

通常情况下，仅执行一次移位算法无法保证解决所有冲突，尤其是当注记密度较大时。因此，本文参考文献[25]，采用迭代策略求解。迭代终止条件包括：(1)Beams 移位方法执行次数($t$)达到设定的阈值($T_s$)，若 $t \geq T_s$，则算法终止；(2)若执行 Beams 移位方法时各注记受力较小，即一定程度上可认为

已解决冲突而无需移位,可基于上一移位过程中各注记的最大受力 $\max(f_{t-1})$ 定义；若 $\max(f_{t-1}) \leq T_f$ ，则算法终止；本文取 $T_f = \varepsilon$ ， $\varepsilon$ 为极小值，实际应用中可由用户设置,如取 $T_f = 0.1d_{\min}$ 。应用 Beams 移位方法进行注记位置配置的流程见算法 1。

**算法 1**：注记位置优化配置的 Beams 算法处理流程
**输入**：点要素集合 $P=\{p_1, p_2, ..., p_m\}$ ，点要素注记对应的最小绑定矩形集合 $R=\{r_1, r_2, ..., r_m\}$ ，点要素与其注记牵引线的连接点集合 $P'=\{p'_1, p'_2, ..., p'_m\}$ ，点 $p_m$ 的牵引线为 $p_m p'_m$ ，迭代次数的终止阈值为 $T_s$ ，注记间或注记与其他要素间的最小距离阈值为 $d_{\min}$ ；
**输出**：移位后的 $R$ 和 $P'$ ；
**初始化**：
　迭代次数 $Step \leftarrow 1$ ,记录上一次迭代处理的最大注记移位量为 $Cache\_max(f_{t-1}) \leftarrow 0$ ；
**Do**
　依据 $R$ 的中心构建三角网并调整；
　计算注记的移位向量；
　基于 Beams 移位方法计算移位向量,最大移位量为 $\max(f_t)$, $Cache\_max(f_{t-1}) \leftarrow Cache\_max(f_t)$ ；
　依据移位向量更新 $R$ 和 $P'$ 的位置；
　$Step \leftarrow Step+1$ ；
**While** ( $Step \leq T_s$ OR $Cache\_max(f_t) \geq 0.1*d_{\min}$ );
**Return** $R$ 和 $P'$

## 4. 实　验

### 4.1 实验说明

**(1) 实验数据**

采用 MapBox 公司提供的三维地理场景为底图,分别展示两种典型点要素：1) 船舶自动定位(Automatic Identification System,缩写为 AIS)数据,选取 2022 年 10 月 10 日中国近海部分 AIS 开源数据局部,数据由东海船舶轨迹数据(脱敏)编制而成,船舶数量为 47；2)兴趣点(Points of interest,缩写为 POI)数据,选择 MapBox 提供的苏黎世中心城区部分 POI 数据为实验数据,POI 数量为 76。

**(2) 评价指标**

依据章节 2.2 中配置约束评价注记配置结果,包括：1)注记间的冲突个数 $N_{r\text{-}r}$ (距离小于阈值 $d_{\min}$ 或压盖)；2)已有图面要素(包括点要素和已有其它图面注记)与注记间的冲突个数 $N_{r\text{-}p}$ (距离小于 $d_{\min}$ 或压盖)；3) 所有注记移位的总长度 $D_{sum}$ ；4) 邻近注记间相对方向的平均偏差 $A_{ms}$ ,定义见公式(14)[26]。

$$A_{ms} = \frac{\sum_{m=1}^{N} \Delta O_m}{N} \quad (14)$$

$N$ 为章节 3.2 定义的邻近图中边的数量(表示边连接的两注记相邻), $\Delta O_m$ 为移位前后对应边的方向偏差,定义见公式(15)。

$$\Delta O_m = \begin{cases} |O_{ij} - O'_{ij}| & (|O_{ij} - O'_{ij}| < 90) \\ 180 - |O_{ij} - O'_{ij}| & (|O_{ij} - O'_{ij}| \geq 90) \end{cases} \quad (15)$$

$O_{ij}$ 和 $O'_{ij}$ 为移位前后对应边的方向。5)采用算法耗时 $t$ 评价运行效率。

**(3) 实验环境**

基于 MapBox JS 实现本文所提方法,实验平台为一台 CPU 为 AMD Ryzen 7-7840HS w/ Radeon 780M Graphics @3.80 GHz、内存为 16GB 和操作系统为 Windows 11(64 位)的计算机。

**(4) 参数设置**

AIS 数据的注记初始布局参数设置如下：引线长度为 1.5cm、方向为 90°、初始引线连接点位置为注记底端中点、最大注记字号为 12pt。POI 数据的注记初始布局参数设置如下：引线长度为 1cm、方向为 90°、初始引线连接点位置为注记底端中点、最大注记字号为 12pt,实际应用时用户可依据需要调整上述参数。参考地图要素间的最小距离约束设置 $d_{\min}=0.2mm$ ,参考点要素间平均距离 $d_{ave}$ 设置 $T_d$ 设置 $T_d=3d_{ave}$[2]；参考文献[25],Beams 方法的迭代次数与注记数量( $P_{Count}$ )相关,设置 $T_s=P_{count}$ ；同时,为避免 $T_s$ 取值过大或过小,若 $P_{Count} \leq 20$ , $T_s=20$ ；若 $P_{Count} \geq 100$ , $T_s=100$ 。

**(5) 实验对比方法**

本文共与两种方法进行对比：1)不进行注记布局调整(NoP)；2)注记局部调整(LocalP),即参考文献[22]和[23]在局部依据注记冲突(重叠或距离小于阈值)程度依次调整注记位置以避免注记冲突。

### 4.2 结果分析

依据实验说明所述的参数执行 Beams 算法对两个数据集中的初始注记配置结果进行优化,结果见图 6。同时,采用实验说明所述的方法进行对比实验,统计结果见表 1。

由表 1 可知,若注记位置未进行优化配置,

AIS 数据集中注记间存在 26 个冲突，注记与已有图面要素间存在 11 个冲突；POI 数据集中注记间存在 23 个冲突，注记与已有图面要素间存在 26 个冲突。依据本文方法和已有方法对注记位置进行配置后则不存在冲突，如图 6 区域 A 和 B 所示，上述结果说明本文方法和已有方法均能有效避免注记冲突。同时，注记的移动也会破坏原有注记间相对关系，由表 1 可知，依据本文方法对注记进行配置后 AIS 数据集会产生 $10.55°$ 的相对方向偏差，比已有方法小 $8.53°$；POI 数据集则会产生 $10.99°$ 的相对方向偏差，比已有方法小 $1.59°$。这是由于 LocalP 只移动存在冲突的要素以避免注记冲突，会导致部分注记移动距离较大而破坏与邻近要素的空间关系。如图 6 区域 A 和 B 中的注记 MINH PHUNG54 和 MARTA，由于 LocalP 仅移动存在冲突的注记，导致注记 MINH PHUNG54 和 MARTA 均偏移其初始位置较远，且破坏了与邻近要素间的相对关系，而本文方法则未出现上述问题。上述结果说明本文方法相比已有方法能更好保持注记间相对关系，这是由于本文方法采用邻近图作为整体框架控制注记移位过程中的空间关系变化。但是，由于 Beams 算法采用全局优化策略，在效率上相比已有局部调整方法存在差距，如 AIS 和 POI 数据集算法耗时分别为 7.28 秒和 10.34 秒，而启发式方法仅为 1.02 秒和 1.72 秒。但是，本文算法可通过划分子群以提高效率，将在章节 5.1 进行详细讨论，AIS 和 POI 数据集在划分后算法耗时可分别减少为 0.42 秒和 1.12 秒。同时，依据本文方法对注记进行配置后 AIS 数据集的注记移动总长度为 13.12cm，比已有方法大 0.51cm；POI 数据集的注记移动总长度为 6.60cm，则比已有方法小 0.84cm。上述结果表明两个方法在注记移位总长度上均未表现出明显优势。

表 1. 注记配置结果评价

|  |  | $N_{r-r}$ | $N_{r-p}$ | $A_{ms}$ | $D_{sum}$ | $t$ |
|---|---|---|---|---|---|---|
| AIS 数据 | NoP | 26 | 11 | 0 | 0 | ---- |
|  | LocalP | 0 | 0 | 19.08 | **12.61** | **1.02s** |
|  | ours | 0 | 0 | **10.55** | 13.12 | 7.28s |
| POIs 数据 | NoP | 23 | 26 | 0 | 0 | ---- |
|  | LocalP | 0 | 0 | 12.58 | 7.44 | **1.72s** |
|  | ours | 0 | 0 | **10.99** | **6.60** | 10.34s |

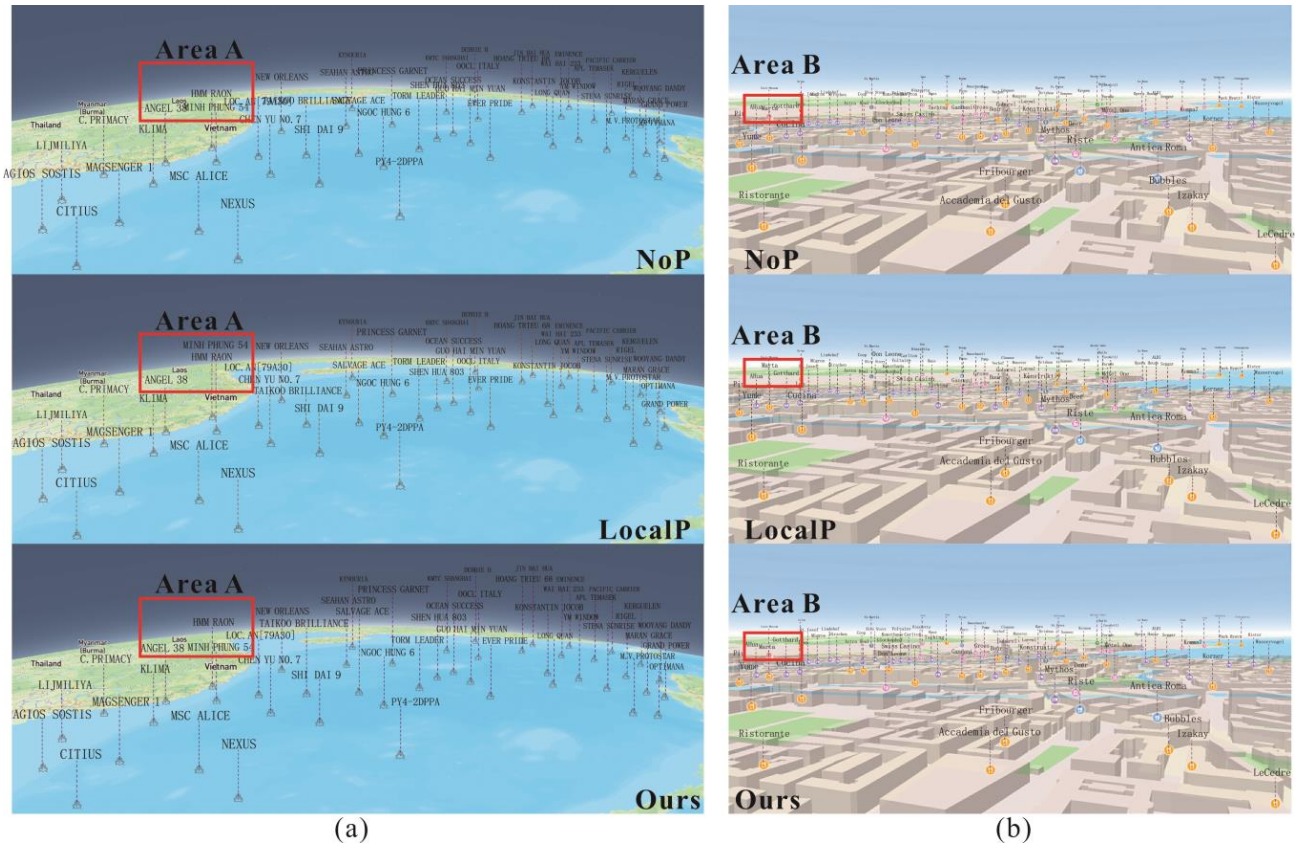

图 6.注记配置结果。（a）AIS 数据集注记配置结果；（b）POI 数据集注记配置结果。

## 5. 讨　论

### 5.1 效率分析

（1）效率分析

为分析本文算法效率的适用性，选取了分别包含 10、30、60、120、200、500 和 1000 个 AIS 注记点的视图（Data A-G）进行注记配置。保持相同的参数设置，实验结果见表 2。由表 2 可知，当注记数量大于 200 时应用本文方法仍然存在 21 个冲突，即移动注记已无法避免冲突。这是由于视图图面空间有限所致，即同一视图无法显示过多注记。同时，由表 2 可知，算法迭代次数（$Step$）与注记点个数无明显关系；而算法耗时（$t$）会随着注记点个数的增大而明显增大，当点总数量超过 60 时，算法耗时超过 8 秒，这是由于 Beams 算法涉及较多的矩阵计算，其计算效率会随数据规模扩大而降低。综合考虑算法效率和图面空间限制，本文方法在注记量小于 60 时，一定程度上可满足三维场景注记即时配置的要求；当注记数量较多时，可能需要先对注记数量进行删除以避免图面要素过多。

表 2. 效率分析（$Step$ 表示算法迭代次数）

| Data | A | B | C | D | E | F | G |
| --- | --- | --- | --- | --- | --- | --- | --- |
| $Step$ | 4 | 18 | 47 | 66 | 100 | -- | -- |
| $t$ | 0.33 | 1.23 | 8.1 | 207.1 | 891.2 | -- | -- |
| $N_{r-r}$ | 0 | 0 | 0 | 0 | 21 | -- | -- |

（2）加速策略

由效率分析结果可知，当图面要素超过 60 时，本方法配置效率大于 8 秒。为满足注记的快速配置要求，仍需要优化本文方法以提高效率。参考文献[16]和[17]采用聚类方法将注记群整体划分为多个子群求解，本研究亦可参考上述策略提高效率。其中，基于最小生成树(Minimum Spanning Tree, 缩写为 MST)进行空间聚类是较常采用的做法，因此，本文亦基于注记间最短距离构建 MST，通过依次删除 MST 中的长边将注记划分为多个子群分别进行处理，子群划分的终止条件为子群中注记的数量 $T_{num}$[27]。分别取 $T_{num}$ 为 10、20 和 30 对 AIS 数据集进行划分，对划分后的子群分别利用 Beams 方法求解并统计算法耗时($t$)，统计结果见表 3。由表 3 可知，对注记划分为子群处理后可以有效提高算法效率，子群划分越小，效率提高越明显，当 $T_{num}$ 分别为 10、20、30 和 ∞ 时，算法耗时分别为 0.42s、1.33s、4.24s 和 7.28s；同时，随着子群的划分，注记的移位总长度和邻近注记间相对方向的平均偏差也会增大；但是，平均方向偏差仍都小于已有方法的 12.58°，分别小 7.93°，8.24° 和 8.15° 和 8.53°。

表 3. 效率分析

| | $N_{r-r}$ | $N_{r-p}$ | $A_{ms}$ | $D_{sum}$ | $t$ |
| --- | --- | --- | --- | --- | --- |
| $T_{num}$=10 | 0 | 0 | 11.15 | 13.35 | 0.42 |
| $T_{num}$=20 | 0 | 0 | 10.84 | 13.24 | 1.33 |
| $T_{num}$=30 | 0 | 0 | 10.93 | 13.18 | 4.24 |
| $T_{num}$ = ∞ | 0 | 0 | 10.55 | 13.12 | 7.28 |

### 5.2 引线类型调整

实际应用中注记的引线类型有时会依据用户需要进行调整（见章节 2.1 引线类型的划分），本文方法亦可灵活扩展适应不同引线类型的注记配置。

**(1) 引线方向固定连接点位置固定**

引线方向固定连接点位置固定即注记只能沿引线的固定方向移动，如设置引线方向为 90°，则注记只能沿竖直方向上下移动。因此，若生成的注记在水平方向上超出屏幕边缘，则仅靠移动注记无法避免此类问题，需对注记及其对应点要素做删除处理，如图 6 中的注记 AGIOS SOSTIS。同时，在受力计算时，需要以引线方向为主轴对注记的受力进行分解，仅保留注记在固定方向上的受力；另外，Beams 算法迭代求解时，更新注记的位置也需要将注记的移位向量进行分解，仅在固定方向上移动注记。依据上述设置对 AIS 数据中的注记位置进行配置，结果见图 7(a)和表 4。由图 7(a)和表 4 可知，本文方法亦可依据引线方向固定连接点位置固定设置对注记位置进行配置，且生成的结果无冲突，满足注记配置要求；同时，对比图 6 和图 7(a)可知，由于图 7(a)固定了连接点位置，因此注记的移动空间相对较小，尽管其总体移位距离减小了 2.17cm，但是其邻近注记间的相对方向相对于其对应点要素之间的相对方向变化则增大了 8.71°。实际应用中，用户若想保持较小的邻近注记间相对方向变化，建议选择引线方向固定连接点位置不固定的设置；反之，可选择引线方向固定连接点位置固定的设置。

**(2) 引线方向不固定连接点位置固定**

引线方向不固定连接点位置固定即注记可在

任意方向上移动，故相对引线方向固定而连接点位置不固定的设置，其没有引线连接点超出注记边缘的约束，即不用考虑章节 3.3 中引线连接点与注记之间的吸引力。依据上述设置对 POIs 数据中的注记位置进行配置，结果见图 7(b)和表 4。由图 7(b)和表 4 可知，本文方法亦可依据引线方向不固定连接点位置固定设置对注记位置进行配置，且生成的结果无冲突，满足注记配置要求；同时，对比图 6 和图 7(b)可知，由于图 7(b)不固定引线方向，注记在位置移动上具有更大的自由度，其总体移位距离和邻近注记间的相对方向偏差分别减少了 0.10cm 和 0.29°。但是，由于引线方向不固定，其亦可能带来视觉上的混乱，如图 7(b)中的区域 A。实际应用中，用户可依据注记实际配置的视觉效果选择合理的引线类型或结合多种引线类型实现注记配置。

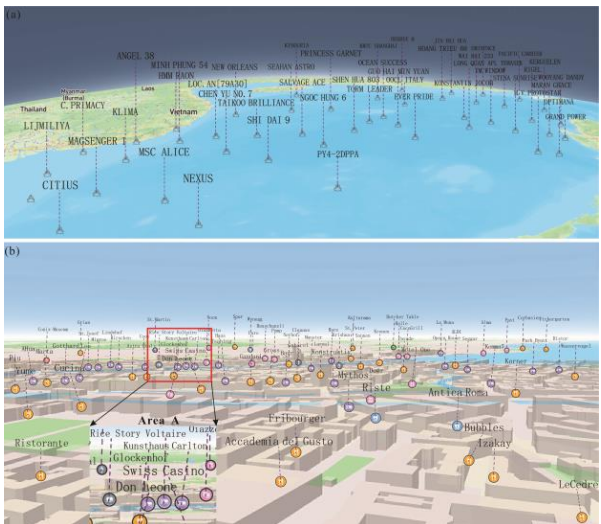

图 7. (a)AIS 数据注记配置结果(固定方向固定连接点位置)；(b)POIs 数据注记配置结果(不固定方向固定连接点位置)。

表 4. 引线类型分析

|  | $N_{r-r}$ | $N_{r-p}$ | $A_{ms}$ | $D_{sum}$ |
|---|---|---|---|---|
| AIS 数据(固定方向不固定连接点) | 0 | 0 | 10.55 | 13.12 |
| AIS 数据(固定方向固定连接点) | 0 | 0 | 19.26 | 10.95 |
| POIs 数据(固定方向固定连接点) | 0 | 0 | 10.99 | 6.60 |
| POIs 数据(不固定方向固定连接点) | 0 | 0 | 10.89 | 6.31 |

### 5.3 邻近图影响

本文利用调整后的 DT 表达注记间空间关系，而描述地图目标间空间关系的邻近图还有 MST 等[28]。其中，MST$\subseteq$DT。因此，本文基于 MST 表达注记间空间关系，并利用本文方法对注记位置进行配置，结果见图 8。统计 Beams 移位方法迭代总次数($Step$)和运行时间 $t$ 分析算法效率，并依据章节 4.1 评价指标分析不同邻近图对于结果影响。由图 6 和图 8 可知，利用 MST 表达注记间空间关系亦能生成满足条件的结果，即注记间或注记与已有图面要素无冲突。同时，由于 MST$\subseteq$DT，即 MST 约束了更少的邻近关系，其算法迭代次数($Step$) 会减少 6 次，算法运行时间($t$)会减少 0.71 s，效率更高；另外，移动的总距离会减小 33.1%，为 4.34cm。但是，由于 MST 约束了更少的邻近关系，邻近注记间的相对方向偏差也会上升，增加 0.65°。实际应用中用户若对效率有更高需要，可选择 MST 构建邻近图；若用户需要较好保持注记间的方向关系，则可选择 DT 构建邻近图。

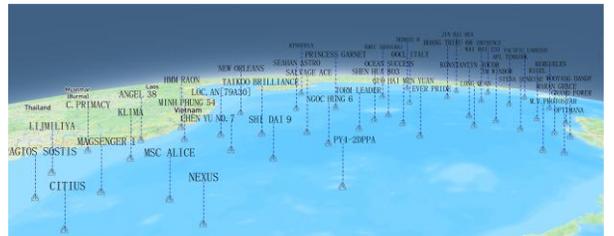

图 8. 基于 MST 的 AIS 数据注记位置配置结果

表 5. 邻近图影响分析

|  | $N_{r-r}$ | $N_{r-p}$ | $A_{ms}$ | $D_{sum}$ | $Step$ | $t$ |
|---|---|---|---|---|---|---|
| MST | 0 | 0 | 11.10 | 8.78 | 32 | 6.57s |
| DT | 0 | 0 | 10.55 | 13.12 | 38 | 7.28s |

### 5.4 字符类型影响

实际应用中注记会采用不同语言，本文实验涉及的 AIS 数据和 POI 数据注记均为英文，故选择 MapBox 提供的北京市王府井地区 POI 数据验证本文方法中文注记支持能力，POI 数量为 88。注记初始布局配置参数如下：引线长度为 1cm、方向为 90、初始引线连接点位置为注记底端中点、最大注记字号为 12pt。实验结果见图 9，统计分析结果见表 6。由图 9 和表 6 可知，若注记位置未进行优化配置时注记间存在 12 个冲突，注记与已有图面要素间存在 18 个冲突，依据本文方法对注记位置进行配置后注记冲突为 0。同时，依据本文方法对注记位置进行配置后，邻近注记间的相对方向相对其对应点要素之间的相对方向偏差为 4.53°，注记移位总长度为 6.12cm。上述结果表明本文方法针对中文注记也能实现无冲突的注记位置配置，且相对方向偏差较小。

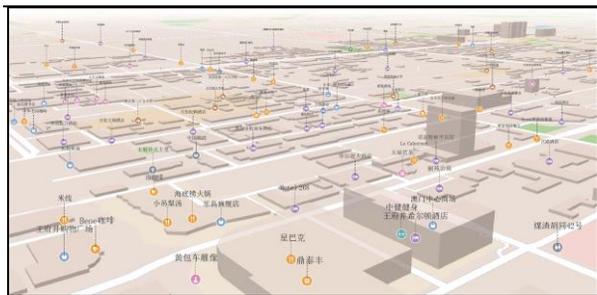

图 9. 中文注记配置结果

表 6. 注记配置结果评价

| | $N_{r-r}$ | $N_{r-p}$ | $A_{ms}$ | $D_{sum}$ |
| --- | --- | --- | --- | --- |
| 初始配置结果 | 12 | 18 | 0 | 0 |
| 本文方法配置结果 | 0 | 0 | 4.53 | 6.12 |

## 6　结　语

为有效实现三维地理场景下带引线点要素注记的位置配置，本文面向二维屏幕显示需要将该问题建模为地图面目标移位问题，应用经典的 Beams 移位方法迭代计算满足配置约束的注记布局。鉴于 Beams 方法利用邻近图表达注记间空间关系，并将该邻近图作为整体框架控制注记移位过程中的空间关系变化，本文方法能更好保持注记间的相对关系，实验结果也有效证明了上述结论。同时，本文也讨论了结合聚类来提高注记配置效率的策略和分析了不同邻近图构建策略对实验结果的影响。但是，本文主要考虑固定视角下的注记配置质量，未有效考虑三维场景中注记随视角变换而动态调整的一致性和配置效率。结合空间索引等研究高效的矩阵运算策略和综合考虑用户交互过程设计适时的注记配置方法是未来研究的重点。

## 参考文献(**References**):

[1] 廖克. 中国地图学发展的回顾与展望[J]. 测绘学报, 2017, 46(10): 1517-1525. (Liao K. Retrospect and Prospect of the Development of Chinese Cartography[J]. Acta Geodaetica et Cartographica Sinica, 2017, 46(10): 1517-1525.)

[2] 何宗宜, 宋鹰, 李连营. 地图学[M]. 武汉: 武汉大学出版社, 2016. (He Z, Song Y, Li L. Cartography[M]. Wuhan: Wuhan University Press, 2016.)

[3] 杨乃, 郭庆胜, 杨族桥. 3 维面状要素注记的自动配置研究[J]. 测绘科学技术学报, 2009, 26(06):449-453. (Yang N, Guo Q, Yang Z. Research on Automatic Placement of Label in3D Area Feature[J]. Journal of Geomatics Science and Technology, 2009, 26(06):449-453.)

[4] Christensen J, Marks J, Shieber S. An empirical study of algorithms for point-feature label placement[J]. ACM Transactions on Graphics (TOG), 1995, 14(3): 203-232.

[5] Edmondson S, Christensen J, Marks J, et al. A general cartographic labelling algorithm[J]. Cartographica: The International Journal for Geographic Information and Geovisualization, 1996, 33(4): 13-24.

[6] 郑春燕, 郭庆胜, 刘小利. 基于禁忌搜索算法的点状要素注记的自动配置[J]. 武汉大学学报(信息科学版), 2006, 31(5):428-431. (Zheng C, Guo Q, Liu X. Automatic Placement of Point Annotation Based on Tabu Search[J]. Geomatics and Information Science of Wuhan University, 2006, 31(5):428-431.)

[7] 罗广祥, 徐斌. 基于 Voronoi 图的点状要素注记自动配置[J]. 长安大学学报(地球科学版), 2003(2): 63-65, 69. (Luo G, Xu B. The study on automatic name placement around point features based on Voronoi[J]．Journal of Chang' an University (Earth Science Edition), 2003(2): 63-65, 69.)

[8] 胡冯伟, 乔俊军, 陈张健等. 改进的地图制图点注记配置探测信息模型[J]. 测绘学报, 2021, 50(1):132-141. (Hu F, Qiao J, Chen Z, et al. An improved detecting information model of point annotation labelling in cartography[J]. Acta Geodaetica et Cartographica Sinica, 2021, 50(1):132-141.)

[9] 张志军. 基于规则引擎的地图注记自动配置方法研究[D]. 武汉大学, 2015. Zhang Z. Research on automatic label placement based on rules-engine[D]. Wuhan University, 2015.

[10] Lu F, Deng J, Li S, et al. (2019). A hybrid of differential evolution and genetic algorithm for the Multiple Geographical Feature Label Placement Problem. ISPRS International Journal of Geo-Information, 8(5), 237.

[11] 彭珊鸽, 宋鹰, 吴凡. 基于蚁群算法的点状注记智能化配置[J]．测绘科学, 2007, 32(5):80-81. (Peng S, Song Y, Wu F. The research of intelligent point-feature cartographic label placement base on ant colony algorithm [J]. Science of Surveying and Mapping, 2007, 32(5):80-81.)

[12] 王立, 郑昊. 粒子群遗传混合算法在点状注记配置中的应用[J]．计算机与现代化, 2012(10): 30-33, 37. (Wang L, Zhen H. A hybrid algorithm of PSO and GA for automatic placement of point annotation[J]. Computer and Modernization, 2012(10): 30-33, 37.)

[13] 李娟, 朱勤东. 一种顾及道路影响的点要素注记配置遗传禁忌搜索算法[J].测绘通报, 2019(2): 80-85. (Li J, Zhu Q. A genetic taboo search algorithm for point feature label placement considering the constrain of road network[J]. Bulletin of Surveying and Mapping, 2019(2): 80-85.)

[14] Araujo E, Chaves A, Lorena L. Improving the Clustering Search heuristic: An application to cartographic labeling[J]. Applied Soft Computing, 2019, 77: 261-273.

[15] Lessani M N, Deng J, Guo Z. A Novel Parallel Algorithm with Map Segmentation for Multiple Geographical Feature Label Placement Problem[J]. ISPRS International Journal of Geo-Information, 2021, 10(12): 826.

[16] 周鑫鑫, 孙在宏, 吴长彬等. 地图点要素注记自动配置中聚类分组的蚁群算法应用[J]．地球信息科学学报，2015, 17(8):902-908. (Zhou X, Sun Z, Wu C, et al．Automatic label placement of point feature: using ant colony algorithm based on group clustering[J]. Journal of Geo-Information Science, 2015, 17(8):902-908.)

[17] 曹闻, 彭斐琳, 童晓冲等. 顾及空间分布与注记相关性的点要素注记配置算法[J]. 测绘学报, 2022, 51(2): 301-311. (Cao W, Peng F, Tong X, et al. A point-feature label placement algorithm considering spatial distribution and label correlation[J]. Acta Geodaetica et Cartographica Sinica, 2022, 51(2): 301-311.

[18] 周鑫鑫, 吴长彬, 孙在宏等.小规模地理场景中点要素三维注记优化配置算法[J]. 测绘学报, 2016, 45(12):14761-1484. (Zhou X, Wu C, Sun Z, et al. A 3D Annotation Optimal Place-

ment Algorithm for the Point Features in the Small Scale Geographic Scene[J]. Acta Geodaetica et Cartographica Sinica, 2016, 45(12): 1476-1484.)

[19] She J, Li X, Liu J, et al. A building label placement method for 3D visualizations based on candidate label evaluation and selection[J]. International Journal of Geographical Information Science, 2019, 33(10): 2033-2054.

[20] Huang Y, Kong D, Zhang Y. Study on dynamic labeling of building in a 3D virtual city. Electrical & Electronics Engineering, 2012, 156-162.

[21] Chen C, Zhang L, Ma J, et al. Adaptive multi-resolution labeling in virtual landscapes[J]. International Journal of Geographical Information Science, 2010, 24(6): 949-964.

[22] Gemsa A, Haunert J H, Nöllenburg M. Multirow boundary-labeling algorithms for panorama images[J]. ACM Transactions on Spatial Algorithms and Systems (TSAS), 2015, 1(1): 1-30.

[23] 郝天浩, 林志勇. 三维室内空间注记优化配置方法[J]. 地理空间信息, 2022, 20(1): 27-31+51+6-7. (Hao T, Lin Z. Optimal 3D Indoor Space Annotation Allocation Method[J]. Geo-Spatial Information, 2022, 20(1): 27-31+51+6-7.)

[24] Tatzgern M, Kalkofen D, Grasset R, et al. Hedgehog labeling: View management techniques for external labels in 3D space. In 2014 IEEE Virtual Reality (VR)[J]. 2014.

[25] 刘远刚. 基于能量最小化原理的地图要素移位算法研究与改进[D]. 武汉大学, 2017. (Liu Y. Research and Improvement of Cartographic Displacement Algorithms Based on Energy Minimization Principles[D]. Wuhan University, 2017.)

[26] Wei Z, Ding S, Xu W, et al. Elastic beam algorithm for generating circular cartograms[J]. Cartography and Geographic Information Science, 2023: 1-14.

[27] Wei Z, Guo Q, Wang L, et al. On the spatial distribution of buildings for map generalization[J]. Cartography and Geographic Information Science, 2018, 45(6): 539-555.

[28] 郭庆胜, 魏智威, 王勇, 王琳. 特征分类与邻近图相结合的建筑物群空间分布特征提取方法[J]. 测绘学报, 2017, 46(5): 631-638. (Guo Q, Wei Z, Wang Y, et al. The method of extracting spatial distribution characteristics of buildings combined with feature classification and proximity graph[J]. Acta Geodaetica et Cartographica Sinica, 2017, 46(5): 631-638.)